\documentclass[10pt]{article}

\usepackage[utf8]{inputenc} 
\usepackage[T1]{fontenc}    
\usepackage{hyperref}       
\usepackage{url}            
\usepackage{booktabs}       
\usepackage{amsfonts}       
\usepackage{nicefrac}       
\usepackage{microtype}      
\usepackage{multicol}
\usepackage{algorithm}
\usepackage{algorithmic}
\usepackage[numbers]{natbib}

\usepackage{amsmath, amsthm, amssymb, multirow, paralist}

\usepackage{graphicx}
\usepackage{epsfig}
\usepackage{comment}
\usepackage{color}
\usepackage{dsfont}
\usepackage{subcaption}
\usepackage{wrapfig}
\usepackage{amsmath}
\usepackage{amssymb}
\usepackage{multirow} 
\usepackage{enumerate}
\usepackage{caption}
\usepackage[body={6.5in,8in}]{geometry}

\title{SemiNLL: A Framework of Noisy-Label Learning by Semi-Supervised Learning}
\date{}

\author{
Zhuowei Wang$^{1}$, Jing Jiang$^{1}$, Bo Han$^{2,4}$, Lei Feng$^{3}$,\\ Bo An$^{3}$, Gang Niu$^{4}$, Guodong Long$^{1}$\\
\\
  $^1$University of Technology Sydney, Australia\\
  $^2$Hong Kong Baptist University, HKSAR, China\\
  $^3$Nanyang Technological University, Singapore\\
  $^4$RIKEN Center for Advanced Intelligence Project, Japan\\
  
  \texttt{\{zhuowei.wang\}@student.uts.edu.au},
  \texttt{\{jing.jiang, guodong.long\}@uts.edu.au},\\
  \texttt{\{feng0093, boan\}@ntu.edu.sg},  

  \texttt{\{bo.han, gang.niu\}@riken.jp}, 
}

\begin{document}

\maketitle

\begin{abstract}
Deep learning with noisy labels is a challenging task.
Recent prominent methods that build on a specific sample selection (SS) strategy and a specific semi-supervised learning (SSL) model achieved state-of-the-art performance.
Intuitively, better performance could be achieved if stronger SS strategies and SSL models are employed. Following this intuition, one might easily derive various effective noisy-label learning methods using different combinations of SS strategies and SSL models, which is, however, reinventing the wheel in essence.
To prevent this problem, we propose \emph{SemiNLL}, a versatile framework that combines SS strategies and SSL models in an end-to-end manner. Our framework can absorb various SS strategies and SSL backbones, utilizing their power to achieve promising performance. We also instantiate our framework with different combinations, which set the new state of the art on benchmark-simulated and real-world datasets with noisy labels.
\end{abstract}
\section{Introduction}
Deep Neural Networks (DNNs) have achieved great success in different computer vision problems, e.g., image classification~\cite{krizhevsky2017imagenet}, detection~\cite{ren2016faster}, and semantic segmentation~\cite{long2015fully}. Such success is demanding for large datasets with clean human-annotated labels. However, it is costly and time-consuming to correctly label massive images for building a large-scale dataset like ImageNet~\cite{imagenet_cvpr09}. Some common and less expensive ways to collect large datasets are through online search engines~\cite{schroff2010harvesting} or crowdsourcing~\cite{yu2018learning}, which would, unfortunately, bring noisy labels to the collected datasets. Besides, an in-depth study~\cite{zhang2016understanding} showed that deep learning with noisy labels can lead to severe performance deterioration. Thus, it is crucial to alleviate the negative effects caused by noisy labels for training DNNs.

A typical strategy is to conduct {\em sample selection} (SS) and to train DNNs with selected samples~\cite{han2018co,jiang2018mentornet,song2019selfie,wei2020combating,yu2019does}. Since DNNs tend to learn simple patterns first before fitting noisy samples~\cite{arpit2017closer}, many studies utilize the small-loss trick, where the samples with smaller losses are taken as clean ones. For example, {\em Co-teaching}~\cite{han2018co} leverages two networks to select small-loss samples within each mini-batch for training each other. Later, \citet{yu2019does} pointed out the importance of the disagreement between two networks and proposed {\em Co-teaching+}, which updates the two networks using the data on which the two networks hold different predictions. By contrast, {\em JoCoR}~\cite{wei2020combating} proposes to reduce the diversity between two networks by training them simultaneously with a joint loss calculated from the selected small-loss samples.
Although these methods have achieved satisfactory performance by training with selected small-loss samples, they simply discard other large-loss samples which may contain potentially useful information for the training process.

To make full use of all given samples, a prominent strategy is to 
consider selected samples as labeled ``clean'' data and other samples as unlabeled data, and to perform {\em semi-supervised learning} (SSL)~\cite{arazo2020pseudo,berthelot2019mixmatch,laine2016temporal,tarvainen2017mean}. Following this strategy, {\em SELF}~\cite{nguyen2019self} detects clean samples by progressively removing noisy samples whose self-ensemble predictions of the model do not match the given labels in each iteration. With the selected labeled and unlabeled data, the problem becomes an SSL problem, and a {\em Mean-Teacher} model \cite{tarvainen2017mean} can be trained.
Another recent method, {\em DivideMix}~\cite{li2020DivideMix}, leverages Gaussian Mixture Model (GMM)~\cite{permuter2006study} to distinguish clean (labeled) and noisy (unlabeled) data, and then uses a strong SSL backbone called {\em MixMatch}~\cite{berthelot2019mixmatch}. {\em DivideMix} achieves state-of-the-art results across different benchmark datasets. 

As shown above, both methods rely on a specific SS strategy and a specific SSL model. The two components play a vitally important role for combating label noise, and stronger components are expected to achieve better performance.
This motivates us to investigate a general algorithmic framework that can leverage various SS strategies and SSL models. 
In this paper, we propose {\em SemiNLL}, which is a versatile framework to bridge the gap between SSL and {\em noisy-label learning} (NLL). 
Our framework can absorb various SS strategies and SSL backbones, utilizing their power to achieve promising performance. Guided by our framework, one can easily instantiate a specific learning algorithm for NLL, by specifying a commonly used SSL backbone with an SS strategy.
The key contributions of our paper can be summarized as follows:
\begin{itemize}
\item To avoid reinventing the wheel for NLL using SSL algorithms, we propose a versatile framework that can absorb various SS strategies and SSL algorithms. Our framework is advantageous since better performance would be achieved if stronger components (including the ones proposed in the future) are used. 
\item To instantiate our framework, we propose {\em DivideMix+} by replacing the epoch-level selection strategy of {\em DivideMix}~\cite{li2020DivideMix} with a mini-batch level one. We also propose {\em GPL}, another instantiation of our framework that leverages a two-component {\em \textbf{G}aussian mixture model}~\cite{li2020DivideMix,permuter2006study} to select labeled (unlabeled) data and uses {\em \textbf{P}seudo-\textbf{L}abeling}~\cite{arazo2020pseudo} as the SSL backbone.
\item We conduct extensive experiments on benchmark-simulated and real-world datasets with noisy labels.
Empirical results show that the stronger SS strategies and SSL backbones we use, the better performance {\em SemiNLL} could achieve. In addition, our instantiations, {\em DivideMix+} and {\em GPL}, outperform other state-of-the-art noisy-label learning methods.
\end{itemize}

\section{Related work}
In this section, we briefly review several related aspects on which our framework builds.

\subsection{Learning with noisy labels}
For NLL, most of the existing methods could be roughly categorized into the following groups:

\noindent\textbf{Sample selection}.\quad 
This family of methods regards samples with small loss as ``clean'' and trains the model only on selected clean samples. For example, {\em self-paced MentorNet}~\cite{jiang2018mentornet}, or equivalently {\em self-teaching}, selects small-loss samples and uses them to train the network by itself. 
To alleviate the sample-selection bias in {\em self-teaching}, \citet{han2018co} proposed an algorithm called {\em Co-teaching}~\cite{han2018co}, where two networks choose the next batch of data for each other for training based on the samples with smaller loss values. {\em Co-teaching+}~\cite{yu2019does} bridges the {\em disagreement strategy}~\cite{malach2017decoupling} with {\em Co-teaching}~\cite{han2018co} by updating the networks over data where two networks make different predictions. In contrast, \citet{wei2020combating} leveraged the agreement maximization algorithm~\cite{kumar2010co} by designing a joint loss to train two networks on the same mini-batch data and selected small-loss samples to update the parameters of both networks. The mini-batch SS strategy in our framework belongs to this direction. However, instead of ignoring the large-loss unclean samples, we just discard their labels and exploit the associated images in an SSL setup. 
 
\noindent\textbf{Noise transition estimation}.\quad Another line of NLL is to estimate the noise transition matrix for loss correction~\cite{goldberger2016training,hendrycks2018using,menon2015learning,natarajan2013learning,patrini2017making,wang2020training,xiao2015learning}. \citet{patrini2017making} first estimated the noise transition matrix and trained the network with two different loss corrections. \citet{hendrycks2018using} proposed a loss correction technique that utilizes a small portion of trusted samples to estimate the noise transition matrix. However, the limitation of these methods is that they do not perform well on datasets with a large number of~classes.
 
\noindent\textbf{Other deep learning methods}.\quad Some other interesting and promising directions for NLL include meta-learning~\cite{finn2017model,snell2017prototypical} based, pseudo-label estimation~\cite{lee2013pseudo} based, and robust loss~\cite{feng2020can,ghosh2017robust,ma2020normalized,wang2019symmetric,xu2019l_dmi,zhang2018generalized} based approaches. For meta-learning based approaches, most studies fall into two main categories: training a model that {\em adapts fast to different learning tasks} without overfitting to corrupted labels~\cite{garcia2016noise,li2019learning}, and {\em learning to reweight} loss of each mini-batch to alleviate the adverse effects of corrupted labels~\cite{ren2018learning,shu2019meta,zhang2020distilling}. Pseudo-label estimation based approaches reassign the labels for noisy samples. For example, {\em Joint-Optim}~\cite{tanaka2018joint} corrects labels during training and updates network parameters simultaneously. {\em PENCIL}~\cite{yi2019probabilistic} proposes a probabilistic model, which can update network parameters and reassign labels as label distributions. The family of pseudo-label estimation has a close relationship with semi-supervised learning~\cite{han2019deep,lee2013pseudo,tanaka2018joint,yi2019probabilistic}. Robust loss based approaches focus on designing loss functions that are robust to noisy labels.

\subsection{Semi-supervised learning}
SSL methods leverage unlabeled data to provide additional information for the training model. A line of work is based on the concept of consistency regularization: if a perturbation is given to an unlabeled sample, the model predictions of the same sample should not be too different. Laine and Aila~\cite{laine2016temporal} applied consistency between the output of the current network and the exponential moving average (EMA) of the output from the past epochs.  Instead of averaging the model outputs, Tarvainen and Valpola~\cite{tarvainen2017mean} proposed to update the network on every mini-batch using an EMA of model parameter values. \citet{berthelot2019mixmatch} introduced a holistic approach that well combines {\em MixUp}~\cite{zhang2017MixUp}, entropy minimization, and consistency regularization. Another line of SSL is pseudo-labeling, the objective of which is to generate pseudo-labels for unlabeled samples to enhance the learning process. Recently, \citet{arazo2020pseudo} proposed a method to improve previous pseudo-labeling methods~\cite{iscen2019label,lee2013pseudo} by adding {\em MixUp} augmentation~\cite{zhang2017MixUp} and setting a minimum number of labeled samples per mini-batch to reduce accumulated error of wrong pseudo-labels.

\subsection{Combination of SS and SSL}
Some previous studies that combine a specific SS strategy and a specific SSL backbone could be regarded as special cases in our framework. \citet{ding2018semi} used a pre-trained DNN on the noisy dataset to select labeled samples. In the SSL stage, {\em Temporal Ensembling}~\cite{laine2016temporal} was used to handle labeled and unlabeled data. \citet{nguyen2019self} proposed a progressive noise filtering mechanism based on the {\em Mean-Teacher} model~\cite{tarvainen2017mean} and its self-ensemble prediction. \citet{li2020DivideMix} used a Gaussian Mixture Model (GMM) to divide noisy and clean samples based on their training losses and fitted them into a recent SSL algorithm called {\em MixMatch}~\cite{berthelot2019mixmatch}. 
Each specific component used in these methods has its own pros and cons. This motivates us to propose a versatile framework that can build on a variety of SS strategies and SSL backbones. In other words, all the methods mentioned in this subsection could be taken as instantiations of our framework.
\begin{figure*}[ht]
\centering
\includegraphics[width=0.99\textwidth]{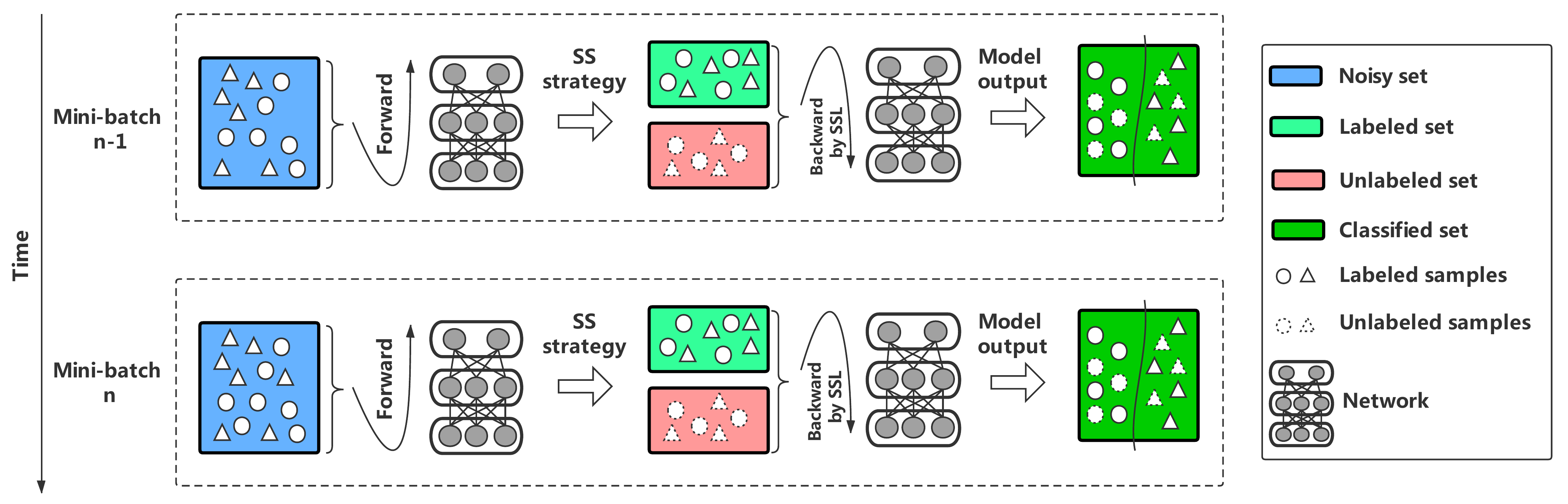}
\caption{The schematic of {\em SemiNLL}. First, each mini-batch of data is forwarded to the network to conduct SS, which divides the original data into the labeled/unlabeled sets. Second, labeled/unlabeled samples are used to train the SSL backbone to produce accurate model output.}

\label{fig:framework}
\end{figure*}

\section{The overview of SemiNLL}

\begin{algorithm}[!t]
\caption{SemiNLL}
\label{alg:seminll}
\begin{algorithmic}[1]
\REQUIRE Network $f_{\theta}$, SS strategy $\textsc{select}$, SSL method $\textsc{semi}$, epoch $T_{\text{max}}$, iteration $I_{\text{max}}$;
\FOR{$t$ = 1,2,\ldots,$T_{\text{max}}$}
    \STATE \textbf{Shuffle} training set ${\mathcal D}_{train}$; 
    \FOR{$n = 1,\ldots,I_{\text{max}}$}
        \STATE \textbf{Fetch} mini-batch $D_n$ from ${\mathcal D}_{train}$;
        \STATE \textbf{Obtain} ${\mathcal{X}}_m, {\mathcal{U}}_m \leftarrow \textsc{select}(D_n, f_{\theta}) $;
        \STATE \textbf{Update} $f_{\theta} \leftarrow \textsc{semi}({\mathcal{X}}_m, {\mathcal{U}}_m, f_{\theta}) $;
    \ENDFOR
\ENDFOR
\ENSURE $f_{\theta}$
\end{algorithmic}
\end{algorithm}
In this section, we present {\em SemiNLL}, a versatile framework of learning with noisy labels by SSL. The idea behind our framework is that we effectively take advantage of the whole training set by trusting the labels of undoubtedly correct samples and utilizing only the image content of potentially corrupted samples. Previous sample selection methods~\cite{han2018co,jiang2018mentornet,wei2020combating,yu2019does} train the network only with selected clean samples, and they discard all potentially corrupted samples to avoid the harmful memorization of DNNs caused by the noisy labels of these samples. In this way, the feature information contained in the associated images might be discarded without being exploited. Our framework, alternatively, makes use of those corrupted samples by ignoring their labels while keeping the associated image content, transforming the NLL problem into an SSL setup. The mechanism of SSL that leverages labeled data to guide the learning of unlabeled data naturally fits well in training the model with the clean and noisy samples divided by our SS strategy. The schematic of our framework is shown in~Figure~\ref{fig:framework}. We first discuss the advantages of the mini-batch SS strategy in our framework and then introduce several SSL backbones used in our framework. 

\subsection{Mini-batch sample selection}\label{subsec:minibatch-separation}
During the SS process, a hazard called confirmation bias \cite{tarvainen2017mean} is worth noting. Since the model is trained using the selected clean (labeled) and noisy (unlabeled) samples, wrongly selected clean samples in this iteration may keep being considered clean ones in the next iteration due to the model overfitting to their labels. Most existing methods~\cite{li2020DivideMix,nguyen2019self} divide the whole training set into the clean/noisy set on an epoch level. In this case, the learned selecting knowledge is incorporated into the SSL phase and will not be updated till the next epoch. Thus, the confirmation bias induced from those wrongly divided samples will accumulate within the whole epoch. To overcome this problem, our mini-batch SS strategy divides each mini-batch of samples into the clean subset ${\mathcal{X}}_m$ and the noisy subset ${\mathcal{U}}_m$ (Line 5 in Algorithm~\ref{alg:seminll}) right before updating the network using SSL backbones. In the next mini-batch, the network can know better to distinguish clean and noisy samples, alleviating the confirmation bias mini-batch by mini-batch.

\subsection{SSL backbones\label{subsec:backbones}}
The mechanism of SSL that uses labeled data to guide the learning of unlabeled data fits well when dealing with clean/noisy data in NLL. The difference lies in an extra procedure, as introduced in Subsection~\ref{subsec:minibatch-separation}, that divides the whole dataset into clean and noisy data. After the SS process, clean samples are considered labeled data and keep their annotated labels. The others are considered noisy samples, and their labels are discarded to be treated as unlabeled ones in SSL backbones. {\em SemiNLL} can build on a variety of SSL algorithms without any modifications to form an end-to-end training scheme for NLL. Concretely, we consider the following representative SSL backbones ranging from weak to strong according to their performance in SSL~tasks:
\begin{enumerate}[(i)]
\setlength\itemsep{0.005em}
    \item {\em Temporal Ensembling}~\cite{laine2016temporal}, where the model uses an exponential moving average (EMA) of label predictions from the past epochs as a target for the unsupervised loss. It enforces consistency of predictions by minimizing the difference between the current outputs and the EMA outputs. 
    \item {\em MixMatch}~\cite{berthelot2019mixmatch}, which is a holistic method that combines {\em MixUp}~\cite{zhang2017MixUp}, entropy minimization, consistency regularization, and other traditional regularization tricks.  It guesses low-entropy labels for augmented unlabeled samples and mixes labeled and unlabeled data using {\em MixUp}~\cite{zhang2017MixUp}.
    \item {\em Pseudo-Labeling}~\cite{arazo2020pseudo}, which learns from unlabeled data by combining soft {\em pseudo-label} generation~\cite{tanaka2018joint} and {\em MixUp} augmentation~\cite{zhang2017MixUp} to reduce confirmation bias in training.
\end{enumerate}

In the next section, we will instantiate our framework by applying specific SS strategies and SSL backbones to the \textsc{select} and \textsc{semi} placeholders in Algorithm~\ref{alg:seminll}.

\section{The instantiations of SemiNLL}
\subsection{Instantiation 1: DivideMix+\label{subsec:dividemix+}}
\begin{figure*}[ht]
\centering
\begin{subfigure}{.7\textwidth}
  \centering
  \includegraphics[width=1\linewidth]{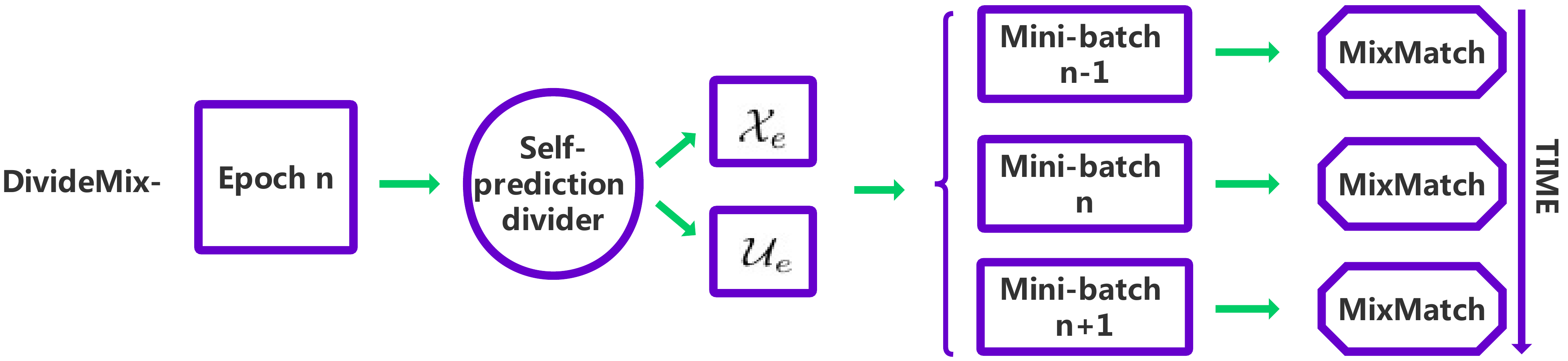}    
  \setlength{\belowcaptionskip}{0.20cm} 
  \caption{DivideMix-}
  \label{fig:a}
\end{subfigure}

\begin{subfigure}{.7\textwidth}
  \centering
  \includegraphics[width=1\linewidth]{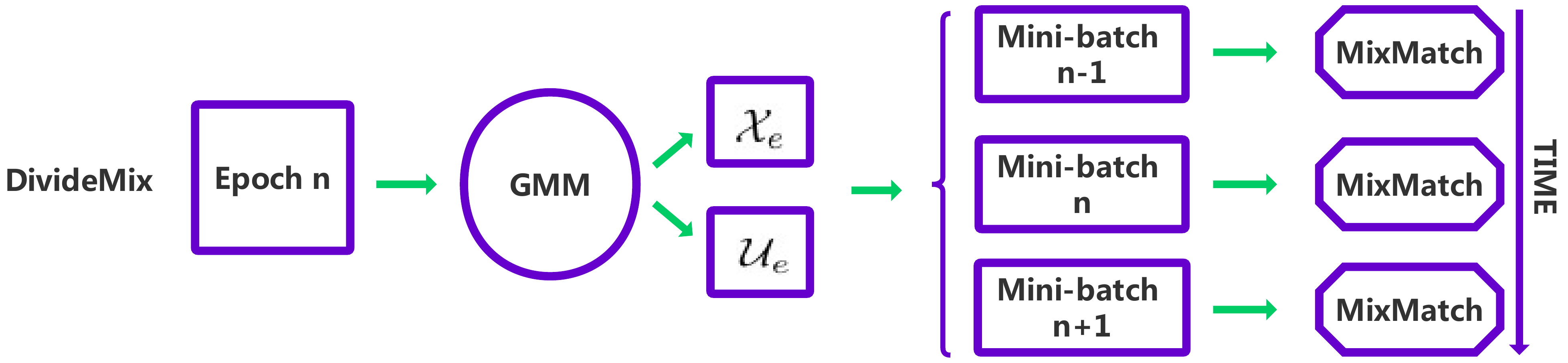} 
      \setlength{\belowcaptionskip}{0.20cm}
  \caption{DivideMix}
  \label{fig:b}
\end{subfigure}

\begin{subfigure}{.7\textwidth}
  \centering
  \includegraphics[width=1\linewidth]{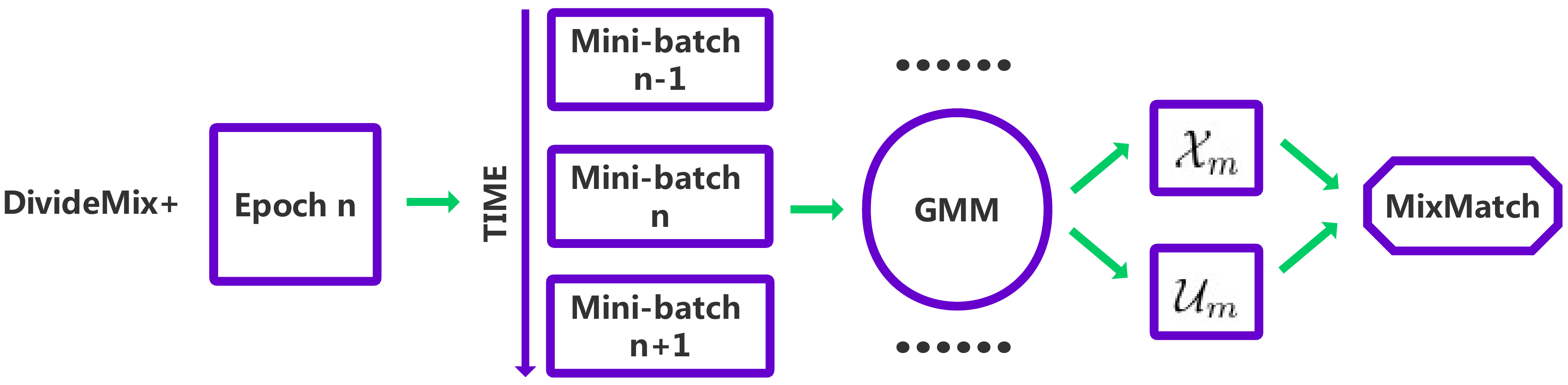}  
  \caption{DivideMix+}
  \label{fig:c}
\end{subfigure}
\caption{Comparisons between: (a) {\em DivideMix-}, (b) {\em DivideMix}, and (c) {\em DivideMix+}. Squares represent data. Circles represent SS strategy. Octagons represent SSL backbone.}

\end{figure*}


In Algorithm~\ref{alg:seminll}, if we (i) specify the \textsc{select} placeholder as a GMM~\cite{permuter2006study}, (ii) specify the \textsc{semi} placeholder as {\em MixMatch}~\cite{berthelot2019mixmatch} mentioned in Subsection~\ref{subsec:backbones}, and (iii) train two independent networks wherein each network selects clean/noisy samples in the SS phase and predicts labels in the SSL phase for the other network, then our framework is instantiated into a mini-batch version of {\em DivideMix}~\cite{li2020DivideMix}. Specifically, during the SS process, {\em DivideMix}~\cite{li2020DivideMix} fits a two-component GMM to the loss $\ell_i$ of each sample using the Expectation-Maximization technique and obtains the posterior probability of a sample being clean or noisy:


\begin{equation}
p\!\left(k\mid\ell_{i}\right)=\frac{p\!\left(k\right)p\!\left(\ell_{i}\mid k\right)}{p\!\left(\ell_{i}\right)},
\end{equation}
where $k=0\left(1\right)$ denotes the clean (noisy) set. 
During the SSL phase, the clean set ${\mathcal{X}}_e$ and the noisy set ${\mathcal{U}}_e$ are fit into an improved {\em MixMatch}~\cite{berthelot2019mixmatch} strategy with label co-refinement and co-guessing. As shown in Figure~\ref{fig:b}, the SS strategy (GMM) of {\em DivideMix}~\cite{li2020DivideMix} is conducted on an epoch level. Since ${\mathcal{X}}_e$ and ${\mathcal{U}}_e$ are updated only once per epoch, the confirmation bias induced from the wrongly divided samples will be accumulated within the whole epoch. However, our mini-batch version, which is called {\em DivideMix+} (Figure~\ref{fig:c}), divides each mini-batch of data into a clean subset ${\mathcal{X}}_m$ and a noisy subset ${\mathcal{U}}_m$, and updates the networks using the SSL backbone right afterwards. In the next mini-batch, the updated networks could better distinguish clean and noisy samples.
\begin{table*}
\centering
\resizebox{0.71\textwidth}{!}{
\setlength{\tabcolsep}{4.25mm}{
\begin{tabular}{c|c|cccc|c}
\toprule
\multirow{3}{*}{Datasets} & \multirow{3}{*}{Method} & \multicolumn{4}{c|}{Symmetric} \\ \cline{3-6}
 &  & \multicolumn{4}{c|}{Noise ratio} & \multirow{1}{*}{Mean}\\
& & 20\% & 40\% & 60\% & 80\%  \\  \midrule
\multirow{7}{*}{MNIST}
& Cross-Entropy & $ 86.16 \pm 0.34  $ & $ 70.39 \pm 0.59  $ & $ 50.35 \pm 0.51  $ & $23.41 \pm 0.96$ & $57.58$ \\
& Coteaching & $ 91.20 \pm 0.03 $ & $90.02 \pm 0.02 $ & $ 83.21 \pm 0.71 $ & $25.33 \pm 0.84 $ & $72.44  $\\
& F-correction & $ 93.93 \pm 0.10  $ & $ 84.30 \pm 0.43  $ & $  65.06 \pm 0.64 $ & $ 29.81 \pm 0.63 $ & $ 68.27 $ \\
& GCE & $ 94.36 \pm 0.11 $ & $93.61 \pm 0.17 $ & $ 92.46 \pm 0.20 $ & $85.04 \pm 0.66 $ & $91.37 $ \\
& M-correction & $ \boldsymbol{97.25 \pm 0.03} $ & $ \underline{96.63 \pm 0.04}$ & $ 95.07 \pm 0.08 $ & $ 86.19 \pm 0.42$  & $ 93.79 $\\
& DivideMix & $ 96.80 \pm 0.08 $ & $96.53 \pm 0.06 $ & $ \underline{96.47 \pm 0.04} $ & $\underline{95.15  \pm 0.25} $ & $  \underline{96.24} $\\

\cmidrule{2-7}
& \textbf{GPL (ours)}& $ 96.67 \pm 0.09 $ & $96.27 \pm 0.08 $ & $ 95.82 \pm 0.09 $ & $94.81 \pm 0.15 $  & $ 95.89 $\\
& \textbf{DivideMix+ (ours)}& $ \underline{96.83 \pm 0.06} $ & $\boldsymbol{96.79 \pm 0.06} $ & $ \boldsymbol{96.69 \pm 0.03} $ & $\boldsymbol{95.91 \pm 0.10} $  & $ \boldsymbol{96.56} $\\
\midrule
\multirow{7}{*}{\begin{tabular}[c]{@{}c@{}}FASHION\\MNIST\\ \end{tabular}} 
& Cross-Entropy & $  90.83 \pm 0.26 $ & $  86.44 \pm 0.11 $ & $ 77.27 \pm 0.56  $ & $  61.84 \pm 1.27 $  & $  79.10$\\
& Coteaching & $ 89.18 \pm 0.32 $ & $89.13 \pm 0.05 $ & $ 80.08 \pm 0.25 $ & $ 60.36 \pm 2.15$ & $  79.69$\\
& F-correction & $ \boldsymbol{ 93.37 \pm 0.17} $ & $ 92.27 \pm 0.06  $  & $ 90.32 \pm 0.30 $ & $85.78 \pm 0.06  $ & $90.43$\\
& GCE & $ \underline{93.35 \pm 0.09} $ & $92.58 \pm 0.11 $ & $ 91.30 \pm 0.20 $ & $88.01  \pm 0.22 $ & $ \underline{91.31} $ \\
& M-correction & $ 93.03 \pm 0.15 $ & $\underline{92.74 \pm 0.42} $ & $ \underline{91.61 \pm 0.02} $ & $85.25 \pm 0.23 $ & $90.66  $ \\
& DivideMix & $ 92.98 \pm 0.17 $ & $92.55 \pm 0.13 $ & $ 91.55 \pm 0.31 $ & $\underline{88.55 \pm 0.24} $ & $ 90.66 $ \\
\cmidrule{2-7}
& \textbf{GPL (ours)} & $ 92.94 \pm 0.20 $ & $91.38 \pm 0.54 $ & $ 89.97 \pm 0.16 $ & $87.14 \pm 0.65  $  & $ 90.36 $\\
& \textbf{DivideMix+ (ours)}& $ 93.20  \pm 0.08 $ & $\boldsymbol{92.89 \pm 0.15} $ & $ \boldsymbol{92.15 \pm 0.16} $ & $ \boldsymbol{88.70 \pm 0.17} $ & $ \boldsymbol{91.74}$ \\
\midrule
\multirow{7}{*}{CIFAR-10}
& Cross-Entropy & $  83.48 \pm 0.17 $ & $  68.49 \pm 0.40 $ & $ 48.65 \pm 0.06  $ & $27.56 \pm 0.43  $  & $  57.05$\\
& Coteaching & $ 67.73 \pm 0.71 $ & $62.83 \pm 0.72 $ & $ 48.81 \pm 0.78 $ & $27.56 \pm 2.71 $ & $ 51.73 $ \\
& F-correction & $ 83.27 \pm 0.04  $ & $ 73.67 \pm 0.30  $ & $ 77.64 \pm 0.11  $ & $63.95 \pm 0.32$ & $74.63$ \\
& GCE & $ 89.72 \pm 0.10 $ & $87.75 \pm 0.05 $ & $ 84.11 \pm 0.26 $ & $ 72.84 \pm 0.30$ & $  83.61$ \\
& M-correction & $ 92.01 \pm 0.40 $ & $90.09 \pm 0.68 $ & $ 85.90 \pm 0.22 $ & $ 70.57 \pm 0.85$ & $  84.64$\\
& DivideMix & $ \underline{94.82 \pm 0.09} $ & $93.95\pm 0.14 $ & $ 92.28 \pm 0.08 $ & $89.30 \pm 0.17 $  & $ 92.59 $\\
\cmidrule{2-7}
& \textbf{GPL (ours)}& $ 94.45 \pm 0.20$ & $\underline{94.00 \pm 0.22}$ & $ \boldsymbol{93.32  \pm 0.10}$ & $\underline{91.76  \pm 0.23}$  & $  \underline{93.38}$\\
& \textbf{DivideMix+ (ours)}& $\boldsymbol{94.84  \pm 0.12}$ & $\boldsymbol{94.03  \pm 0.20}$ & $ \underline{93.08  \pm 0.19}$ & $\boldsymbol{91.91  \pm 0.07} $ & $\boldsymbol{93.47}  $ \\
\midrule
\multirow{6}{*}{CIFAR-100}
& Cross-Entropy & $  60.93 \pm 0.40 $ & $  46.24 \pm 0.74 $ & $ 29.00 \pm 0.38  $ & $  11.42 \pm 0.19$  & $ 36.90 $\\
& F-correction & $ 60.49 \pm 0.29  $ & $ 48.93 \pm 0.21  $ & $ 48.74 \pm 0.41  $ & $22.93 \pm 0.78$ & $45.27$ \\
& GCE & $ 69.20 \pm 0.10$ & $ 65.90 \pm 0.25$ & $57.33 \pm 0.18 $ & $18.19 \pm 1.15 $ & $  52.66$ \\
& M-correction & $ 67.96 \pm 0.17$ & $64.48 \pm 0.76$ & $ 55.37 \pm 0.72$ & $ 24.21 \pm 1.06$  & $ 53.01 $\\
 & DivideMix & $ \underline{73.17  \pm 0.28}$ & $\underline{71.01  \pm 0.16}$ & $ \underline{66.61 \pm 0.18}$ & $ 43.25 \pm 0.82$ & $63.51  $ \\
 \cmidrule{2-7}
 & \textbf{GPL (ours)}& $ 71.24 \pm 0.24$ & $68.89 \pm 0.07$ & $ 65.80 \pm 0.63 $ & $\boldsymbol{59.96  \pm 0.15}$ & $\underline{66.47}  $ \\
& \textbf{DivideMix+ (ours)}& $ \boldsymbol{73.22 \pm 0.21}$ & $ \boldsymbol{71.03  \pm 0.32}$ & $ \boldsymbol{67.52  \pm 0.19}$ & $ \underline{58.07  \pm 0.71}$ & $  \boldsymbol{67.46} $ \\
\bottomrule
\end{tabular}

}
}
\caption{Average test accuracy (\%) and standard deviation (5 runs) of all the methods in various datasets under symmetric label noise. The best accuracy is \textbf{bold-faced}. The second-best accuracy is \underline{underlined}.} \label{table:sym} 
\end{table*}
\subsection{Instantiation 2: GPL}
Intuitively, the choice of stronger SS strategies and SSL models would achieve better performance based on our framework. Thus, we still choose GMM to distinguish clean and noisy samples due to its flexibility in the sharpness of distribution~\cite{li2020DivideMix}. As for the SSL backbone, we choose the strongest {\em Pseudo-Labeling}~\cite{arazo2020pseudo} introduced in Subsection~\ref{subsec:backbones}. We call this instantiation {\em GPL} ({\em \textbf{G}MM} + {\em \textbf{P}seudo-\textbf{L}abeling}). 
Note that we do not train two networks in {\em GPL} as in {\em DivideMix}~\cite{li2020DivideMix} and {\em DivideMix+}. To our understanding, training two networks simultaneously might provide significant improvements in performance. However, this is outside the scope of this paper, since our goal is to demonstrate the versatility of our framework.

\subsection{Self-prediction divider}
Inspired by {\em SELF}~\cite{nguyen2019self}, we introduce the {\em self-prediction divider}, a simple yet effective SS strategy which leverages the information provided by the network's own prediction to distinguish clean and noisy samples. Based on the phenomenon that DNN's predictions tend to be consistent on clean samples and inconsistent on noisy samples in different training iterations, we select the correctly annotated samples via the consistency between the original label set and the model's own predictions. The {\em self-prediction divider} determines potentially clean samples in a mini-batch if the samples' maximal likelihood predictions of the network match their annotated labels. Specifically, the samples are divided into the labeled set only if the model predicts the annotated label to be the correct class with the highest likelihood. The others are considered noisy samples, and their labels will be discarded to be regarded as unlabeled ones in SSL backbones. Compared to previous small-loss SS methods~\cite{han2018co,wei2020combating,yu2019does}, which depend on a known noise ratio to control how many small-loss samples should be selected in each training iteration, {\em self-prediction divider} does not need any additional information to perform SS strategy where the clean subset and the noisy subset are determined by the network itself. Concretely, we instantiate three learning algorithms by combining our {\em \textbf{s}elf-\textbf{p}rediction \textbf{d}ivider (SPD)} with three SSL backbones introduced in Subsection~\ref{subsec:backbones} and denote them as {\em SPD-Temporal Ensembling}, {\em SPD-MixMatch}, and {\em SPD-Pseudo-Labeling}, respectively.
\begin{table*}
\centering
\resizebox{0.71\textwidth}{!}{
\setlength{\tabcolsep}{4.25mm}{
\begin{tabular}{c|c|cccc|c}
\toprule
\multirow{3}{*}{Datasets} & \multirow{3}{*}{Method} & \multicolumn{4}{c|}{Asymmetric} \\ \cline{3-6} 
 &  & \multicolumn{4}{c|}{Noise ratio}  &  \multirow{1}{*}{Mean}\\
& & 10\% & 20\% & 30\% & 40\% \\ \midrule
\multirow{7}{*}{MNIST}
& Cross-Entropy & $  95.78 \pm 0.19 $ & $  91.15 \pm 0.26 $ & $  86.01 \pm 0.25 $ & $79.92 \pm 0.32   $  & $ 88.22 $\\
& Coteaching & $ 90.32 \pm 0.02 $ & $89.03 \pm 0.02 $ & $79.80 \pm 0.27 $ & $64.94 \pm 0.02 $  & $ 81.02 $\\
& F-correction & $ 96.39 \pm 0.04  $ & $ 94.27 \pm 0.21  $ & $  89.33 \pm 0.94 $ & $ 81.61 \pm 0.42 $ & $ 90.40 $ \\
& GCE & $ 94.61 \pm 0.13 $ & $94.43 \pm 0.07 $ & $94.00 \pm 0.12 $ & $93.42 \pm 0.12 $  & $ 94.12 $\\
& M-correction & $ \underline{96.74 \pm 0.03} $ & $ \underline{96.70 \pm 0.10} $ & $ \boldsymbol{96.67  \pm 0.07}$ & $94.85 \pm 0.40 $ & $ 96.24 $ \\
& DivideMix & $96.17 \pm 0.06 $ & $96.11 \pm 0.09 $ & $ 95.88 \pm 0.05$ & $95.83 \pm 0.05 $  & $96.00  $\\
\cmidrule{2-7}
& \textbf{GPL (ours)} & $ \boldsymbol{96.76  \pm 0.04} $ & $\boldsymbol{96.71 \pm 0.03} $ & $96.49 \pm 0.08 $ & $  \underline{96.45 \pm 0.04}$  & $  \boldsymbol{96.60}$ \\
& \textbf{DivideMix+ (ours)} & $ 96.67 \pm 0.04 $ & $96.66 \pm 0.07 $ & $ \underline{96.50  \pm 0.04} $ & $\boldsymbol{96.46  \pm 0.04} $ & $  \underline{96.57} $ \\
\midrule
\multirow{7}{*}{\begin{tabular}[c]{@{}c@{}}FASHION\\MNIST\\ \end{tabular}} 
& Cross-Entropy & $ \underline{93.88 \pm 0.16}  $ & $  92.20 \pm 0.33 $ & $ 90.41 \pm 0.67  $ & $84.56 \pm 0.41$ & $ 90.26 $ \\
& Coteaching & $ 88.01 \pm 0.03$ & $ 78.88 \pm 0.20$ & $ 70.07 \pm 0.38$ & $ 61.97 \pm 0.21$ & $74.73 $\\
& F-correction & $ \boldsymbol{94.17} \pm 0.12  $ & $ \boldsymbol{93.88 \pm 0.10}  $ & $  \boldsymbol{93.50 \pm 0.10}  $ & $  \boldsymbol{93.25 \pm 0.16} $ & $ \boldsymbol{93.7} $ \\
& GCE & $ 93.51  \pm 0.17$ & $ \underline{93.24  \pm 0.14}$ & $\underline{92.21 \pm 0.27}$ & $89.53 \pm 0.53$ & $ 92.12 $\\
& M-correction & $92.11 \pm 0.93 $ & $ 91.26 \pm 1.35 $ & $89.79 \pm 1.28 $ & $89.58 \pm 2.20$ &  $ 90.69 $\\

& DivideMix & $ 91.83 \pm 0.24 $ & $91.09 \pm 0.08 $ & $89.90 \pm 0.26 $ & $ 87.58 \pm 0.26 $ & $ 90.10 $\\
\cmidrule{2-7}
& \textbf{GPL (ours)}& $ 92.52 \pm 0.22 $ & $92.23 \pm 0.09 $ & $  92.15  \pm 0.26$ & $\underline{91.64 \pm 0.31} $& $ \underline{92.14} $ \\
& \textbf{DivideMix+ (ours)}& $  92.56 \pm 0.39 $ & $ 92.25 \pm 0.21 $ & $91.62 \pm 0.08 $ & $  89.67 \pm 0.44 $& $ 91.53 $ \\
\midrule
\multirow{7}{*}{CIFAR-10}
& Cross-Entropy & $ 90.85 \pm 0.06  $ & $  87.23 \pm 0.40$ & $  81.92 \pm 0.32 $ & $  76.23 \pm 0.45 $ & $ 84.06 $\\
& Coteaching & $ 62.85 \pm 2.20 $ & $  61.04 \pm 1.31 $ & $ 54.50 \pm 0.39 $ & $ 51.68 \pm 1.66$& $57.52  $ \\
& F-correction & $ 89.79 \pm 0.33  $ & $  86.79 \pm 0.67 $ & $  83.34 \pm 0.30 $ & $ 76.81 \pm 1.08 $ & $ 84.18 $ \\
& GCE  & $ 90.40 \pm 0.09$ & $89.30 \pm 0.13$ & $86.89 \pm 0.22$ & $ 82.60 \pm 0.17$& $ 87.30 $ \\
& M-correction & $ 92.28 \pm 0.12$ & $92.13 \pm 0.17$ & $91.38 \pm 0.11$ & $ 90.43 \pm 0.23$& $ 91.56 $ \\
& DivideMix & $ 93.61 \pm 0.15$ & $ 92.99 \pm 0.21$ & $ 91.79 \pm 0.36$ & $ 90.57 \pm 0.31$& $92.24  $ \\
\cmidrule{2-7}
& \textbf{GPL (ours)}& $ \boldsymbol{94.32  \pm 0.01}$ & $ \boldsymbol{94.23 \pm 0.07}$ & $\boldsymbol{93.79 \pm 0.06}$ & $ \boldsymbol{93.02 \pm 0.30}$ & $ \boldsymbol{93.84} $\\
& \textbf{DivideMix+ (ours)}& $ \underline{94.27  \pm 0.23}$ & $ \underline{93.92 \pm 0.20}$ & $ \underline{92.82 \pm 0.28}$ & $  \underline{91.91 \pm 0.24}$& $  \underline{93.23} $ \\
\midrule
\multirow{7}{*}{CIFAR-100}
& Cross-Entropy & $  68.58 \pm 0.34 $ & $  68.82 \pm 0.22 $ & $ 53.99 \pm 0.50  $ & $  44.31 \pm 0.23 $& $ 58.93 $ \\
& F-correction & $ 68.87 \pm 0.06  $ & $  64.11 \pm 0.37 $ & $  56.45 \pm 0.59 $ & $ 46.44 \pm 0.50 $ & $58.97  $ \\
& GCE & $70.77 \pm 0.14 $ & $ 69.22 \pm 0.15$ & $64.60 \pm 0.25$ & $51.72 \pm 1.17 $& $  64.08$ \\
& M-correction & $ 69.44 \pm 0.52$ & $67.25 \pm 0.81$ & $63.16 \pm 1.55 $ & $52.90 \pm 1.79 $ & $ 63.19 $\\
 & DivideMix & $ \boldsymbol{74.00  \pm 0.29}$ & $ \underline{73.28  \pm 0.42}$ & $\boldsymbol{72.84  \pm 0.36}$ & $ 54.33 \pm 0.69$& $ 68.61 $ \\
 \cmidrule{2-7}
 & \textbf{GPL (ours)}& $ 71.94 \pm 0.29$ & $71.22 \pm 0.11$ & $ 70.56 \pm 0.23$ & $\boldsymbol{69.84  \pm 0.41} $& $\boldsymbol{ 70.89} $ \\
& \textbf{DivideMix+ (ours)}& $  \underline{73.49 \pm 0.31}$ & $\boldsymbol{73.30  \pm 0.22}$ & $ \underline{72.36 \pm 0.43}$ & $ \underline{55.63  \pm 0.60}$& $  \underline{68.70} $ \\
\bottomrule
\end{tabular}%
}
}
\caption{Average test accuracy (\%) and standard deviation (5 runs) of all the methods in various datasets under asymmetric label noise. The best accuracy is \textbf{bold-faced}. The second-best accuracy is \underline{underlined}.} \label{table:asy} 
\end{table*}

\subsection{Effects of the two components}
This section demonstrates the effects of SS strategies and SSL backbones in our framework. To prove that a more robust SS strategy can boost performance for our framework, we propose {\em DivideMix-} (Figure~\ref{fig:a}) by replacing the GMM in {\em DivideMix}~\cite{li2020DivideMix} with our {\em self-prediction divider} on an epoch level. Since {\em self-prediction divider} is supposed to be weaker than GMM, {\em DivideMix-} is expected to achieve lower performance than {\em DivideMix}~\cite{li2020DivideMix}. To prove the effectiveness of the SSL backbone, we remove it after the SS process and only update the model using the supervised loss calculated from the clean samples. We will give detailed discussions in Subsection~\ref{subsec:effectofstrategy} and Subsection~\ref{subsec:effectofbackbone}.

\section{Experiments}
In this section, we first compare two instantiations of our framework, {\em DivideMix+} and {\em GPL}, with other state-of-the-art methods. 
We also analyze the effects of SS strategies by comparing {\em DivideMix-}, {\em DivideMix}~\cite{li2020DivideMix}, and {\em DivideMix+}, then analyze the effects of SSL backbones by combining three representative SSL methods with our {\em self-prediction divider}. More information of our experiments can be found in supplementary~materials.
\subsection{Experiment setup}
\noindent \textbf{Datasets}. We thoroughly evaluate our proposed {\em DivideMix+} and {\em GPL} on five datasets, including MNIST~\cite{lecun1998gradient-based}, FASHION-MNIST~\cite{xiao2017fashion}, CIFAR-10, CIFAR-100~\cite{krizhevsky2009learning}, and~Clothing1M~\cite{xiao2015learning}. 

MNIST and FASHION-MNIST contain 60K training images and 10K test images of size $28\times28$. CIFAR-10 and CIFAR-100 contain 50K training images and 10K test images of size $32\times32$ with three channels. According to previous studies~\cite{li2020DivideMix,wei2020combating,zhang2018generalized}, we experiment with two types of label noise: symmetric noise and asymmetric noise. Symmetric label noise is produced by changing the original label to all possible labels randomly and uniformly according to the noise ratio. Asymmetric label noise is similar to real-world noise, where labels are flipped to similar classes.

Clothing1M is a large-scale real-world dataset that consists of one million training images from online shopping websites with labels annotated from surrounding texts. The estimated noise ratio is approximately 40\%~\cite{xiao2015learning}.

\begin{table*}
\centering
\resizebox{1.00\textwidth}{!}{
\setlength{\tabcolsep}{3mm}{
\begin{tabular}{c|c|cccc|cccc}
\toprule
\multirow{3}{*}{Datasets} & \multirow{3}{*}{Method} & \multicolumn{4}{c|}{Symmetric} & \multicolumn{4}{c}{Asymmetric} \\ \cline{3-10} 
 &  & \multicolumn{4}{c|}{Noise ratio} & \multicolumn{4}{c}{Noise ratio} \\
& & 20\% & 40\% & 60\% & 80\% & 10\% & 20\% & 30\% & 40\% \\ 
\midrule
\multirow{3}{*}{CIFAR-10}
 & DivideMix- & $ 94.49 \pm 0.02 $ & $93.64 \pm 0.12 $ & $ 91.65 \pm 0.34 $ & $76.61 \pm 1.26$ & $ 93.58 \pm 0.02 $ & $92.87 \pm 0.14 $ & $91.21 \pm 0.21 $ & $ 90.42 \pm 0.23$ \\
 & DivideMix & $ 94.82 \pm 0.09 $ & $93.95\pm 0.14 $ & $ 92.28 \pm 0.08 $ & $89.30 \pm 0.17 $ & $ 93.61 \pm 0.15$ & $ 92.99 \pm 0.21$ & $ 91.79 \pm 0.36$ & $ 90.57 \pm 0.31$ \\
 & DivideMix+ (ours) & $\boldsymbol{94.84  \pm 0.12}$ & $\boldsymbol{94.03  \pm 0.20}$ & $ \boldsymbol{93.08  \pm 0.19}$ & $\boldsymbol{91.91  \pm 0.07} $ & $ \boldsymbol{94.27  \pm 0.23}$ & $ \boldsymbol{93.92 \pm 0.20}$ & $ \boldsymbol{92.82 \pm 0.28}$ & $  \boldsymbol{91.91 \pm 0.24}$\\
\midrule
\multirow{3}{*}{CIFAR-100}
& DivideMix- &  $72.51 \pm 0.32  $ & $ 69.27 \pm 0.46$ & $ 61.13 \pm 0.60 $ & $25.96 \pm 0.78$ & $ 73.62 \pm 0.12 $ & $ 72.32 \pm 0.24$ & $70.64 \pm 0.20 $ & $\boldsymbol{68.04 \pm 1.24} $ \\
& DivideMix & $ 73.17  \pm 0.28$ & $71.01  \pm 0.16$ & $ 66.61 \pm 0.18$ & $ 43.25 \pm 0.82$  & $ \boldsymbol{74.00  \pm 0.29}$ & $ 73.28  \pm 0.42 $ & $\boldsymbol{72.84  \pm 0.36}$ & $ 54.33 \pm 0.69$ \\
& DivideMix+ (ours) & $ \boldsymbol{73.22 \pm 0.21}$ & $ \boldsymbol{71.03  \pm 0.32}$ & $ \boldsymbol{67.52  \pm 0.19}$ & $ \boldsymbol{58.07  \pm 0.71}$ & $  73.49 \pm 0.31$ & $\boldsymbol{73.30  \pm 0.22}$ & $ 72.36 \pm 0.43$ & $ 55.63  \pm 0.60$ \\
\bottomrule
\end{tabular}
}
}
\caption{Test accuracy (\%) of {\em DivideMix-}, {\em DivideMix}, and {\em DivideMix+}.}\label{table:dividemix-}
\end{table*}


\noindent\textbf{Network Structure and Optimizer.}\quad Following previous works~\cite{arazo2020pseudo,li2020DivideMix,wei2020combating,zhang2018generalized}, we use a 2-layer MLP for MNIST, a ResNet-18~\cite{he2016deep} for FASHION-MNIST, the well-known ``13-CNN'' architecture~\cite{tarvainen2017mean} for CIFAR-10 and CIFAR-100, and an 18-layer PreAct Resnet~\cite{he2016identity} for Clothing1M. To ensure a fair comparison between the instantiations of our framework and other methods, we keep the training settings for MNIST, CIFAR-10, CIFAR-100, and Clothing1M as close as possible to {\em DivideMix}~\cite{li2020DivideMix} and FASHION-MNIST close to {\em GCE}~\cite{zhang2018generalized}.

For FASHION-MNIST, the network is trained using stochastic gradient descent (SGD) with 0.9 momentum and a weight decay of $1\times10^{-4}$ for 120 epochs. For MNIST, CIFAR-10, and CIFAR-100, all networks are trained using SGD with 0.9 momentum and a weight decay of $5\times10^{-4}$ for 300 epochs. For Clothing1M, the momentum is 0.9, and the weight decay is 0.001. 


\noindent\textbf{Baselines}.\quad We compare {\em DivideMix+} and {\em GPL} with the following state-of-the-art algorithms and implement all methods by PyTorch on NVIDIA Tesla V100 GPUs.
\begin{enumerate}[(i)]
\setlength\itemsep{0.005em}
    \item {\em Coteaching}~\cite{han2018co}, which trains two networks and cross-updates the parameters of peer networks.
    \item {\em GCE}~\cite{zhang2018generalized}, which uses a theoretically grounded and easy-to-use loss function, the $\mathcal{L}_q$ loss, for NLL.
    \item {\em F-correction}~\cite{patrini2017making}, which corrects the prediction by the label transition matrix. As suggested by the authors, we first train a standard network using the cross-entropy loss to estimate the transition matrix.
    \item {\em M-correction}~\cite{arazo2019unsupervised}, which models clean and noisy samples by fitting a two-component BMM and applies {\em MixUp} data augmentation~\cite{zhang2017MixUp}.
    \item {\em DivideMix}~\cite{li2020DivideMix}, which divides clean and noisy samples by using a GMM on an epoch level and leverages {\em MixMatch}~\cite{berthelot2019mixmatch} as the SSL backbone.
\end{enumerate}


\subsection{Performance Comparison}
The results of all the methods under symmetric and asymmetric noise types on MNIST, FASHION-MNIST, CIFAR-10, and CIFAR-100 are shown in Table~\ref{table:sym} and Table~\ref{table:asy}. The results on Clothing1M are shown in Table~\ref{table:clothing}.

\noindent\textbf{Results on MNIST.}\quad {\em DivideMix+} surpasses {\em DivideMix} across symmetric and asymmetric noise at all noise ratios, showing the effectiveness of the mini-batch SS strategy in our framework. 
{\em M-correction} performs well under low noise ratios. However, in the hardest symmetric 80\% case, {\em DivideMix+} achieves best test accuracy.

\noindent\textbf{Results on FASHION-MNIST.}\quad FASHION-MNIST is quite similar to MNIST but more complicated. {\em DivideMix+} still outperforms {\em DivideMix} on symmetric and asymmetric noise at all noise ratios. 
In the harder asymmetric 40\% noise, {\em DivideMix+} and {\em GPL} outperform the other methods by a large margin. 

\noindent\textbf{Results on CIFAR-10.}\quad {\em DivideMix+} constantly outperforms {\em DivideMix}, especially in the cases with higher noise ratios. 
We believe the reason is that the mini-batch SS strategy used in our framework can better mitigate the confirmation bias induced from wrongly divided samples in more challenging scenarios. 
Overall, {\em GPL} and {\em DivideMix+} surpass the other methods over a large margin, with the latter performing extremely well on asymmetric noise.

\noindent\textbf{Results on CIFAR-100.}\quad 
In most cases, {\em DivideMix+} and {\em DivideMix} achieve higher test accuracy than the other approaches, with {\em DivideMix+} performing better. Specifically, {\em DivideMix+} surpasses {\em DivideMix} by 14.82\% in the hardest symmetric 80\% case.  An interesting phenomenon is that all the approaches suffer from performance deterioration in the asymmetric 40\% cases except {\em GPL}, which significantly outperforms the second-best algorithm over +14\%.

\begin{table*}[ht]
\centering
\resizebox{0.86\textwidth}{!}{
\setlength{\tabcolsep}{4mm}{
\begin{tabular}{l |c c c|c c c}
\toprule
Dataset & \multicolumn{3}{c|}{CIFAR-10}&\multicolumn{3}{c}{CIFAR-100}\\
\midrule
Method/Noise ratio&  20\% & 50\%& 80\% &20\% & 50\%& 80\% \\

\midrule
SPD-Cross-Entropy &   $83.13 \pm 0.16$ &  $79.74 \pm 0.10$& $49.14 \pm 0.15$    & $45.07 \pm 0.55$ &  $35.02 \pm 0.57$& $10.22 \pm 0.10$\\

SPD-Temporal Ensembling &   $83.15 \pm 0.06$  & $80.16 \pm 0.36$ & $49.10 \pm 0.13$    &$46.16 \pm 0.12$    & $ 39.91 \pm 0.60  $ &  $ 12.37 \pm 0.67    $\\

SPD-MixMatch &    $ 93.53 \pm 0.52$  & $90.22 \pm 0.18$ & $88.77 \pm 0.20$     & $72.89 \pm 0.30$   &$68.57 \pm 0.20$  & $33.92 \pm 0.20$\\

SPD-Pseudo-Labeling &    $ \boldsymbol{94.52 \pm 0.06}$    & $\boldsymbol{93.24 \pm 0.36}$ &$ \boldsymbol{90.27 \pm 0.34}$ &  $\boldsymbol{73.84 \pm 0.48}$  & $ \boldsymbol{68.61 \pm 0.40}$ &$ \boldsymbol{55.37 \pm 0.34}$ \\
			
\bottomrule
\end{tabular}
}
}
\setlength{\belowcaptionskip}{-0.43cm} 
\caption{Test accuracy (\%) of the baseline and three SSL backbones integrated into our proposed framework.}\label{table:sslbackbones} 
\end{table*}

\noindent\textbf{Results on Clothing1M.}\quad
To show the robustness of our framework under real-world noisy labels, we demonstrate the effectiveness of {\em DivideMix+} and {\em GPL} on Clothing1M. As shown in Table ~\ref{table:clothing}, the performance of {\em DivideMix+} is better than that of {\em DivideMix} and other methods. 


\subsection{The effects of SS strategies\label{subsec:effectofstrategy}}
To study how SS strategies can affect the performance of our framework, we propose {\em DivideMix-} by replacing the GMM component in {\em DivideMix} with our {\em self-prediction divider} yet maintaining the epoch-level SS strategy for a fair comparison. Due to constraints of space, we only provide the mean value of the results in Table \ref{table:dividemix-}, which can show the overall tendency. Results with mean and standard deviation can be found in supplementary materials. In CIFAR-10, the difference between {\em DivideMix-} and {\em DivideMix} is not obvious in the lower noise ratios. However, in the most difficult symmetric 80\% case, the test accuracy of {\em DivideMix} is +12.69\% higher than {\em DivideMix-}. The difference is even greater in CIFAR-100, showing that GMM is better able to distinguish clean and noisy labels in most cases. An impressive phenomenon to note is that {\em DivideMix-} excels in the asymmetric 40\% case in CIFAR-100, which means the {\em self-prediction divider} performs better in nosier asymmetric cases than GMM. 
The reason is explained in the original paper of {\em DivideMix}~\cite{li2020DivideMix}, that GMM cannot effectively distinguish clean and noisy samples under asymmetric noise with high noise ratio in datasets with a large number of classes. At the same time, the fact that {\em DivideMix+} constantly outperforms {\em DivideMix} in most cases shows that the mini-batch SS strategy in our framework is better than the epoch-level one in {\em DivideMix}.

\subsection{The effects of SSL backbones\label{subsec:effectofbackbone}}
We evaluate the effects of SSL backbones in our framework by combining the {\em self-prediction divider (SPD)} with three different SSL methods and a baseline which only updates the model using the cross-entropy loss calculated from clean samples. We denote them as {\em SPD-Temporal Ensembling}, {\em SPD-MixMatch}, {\em SPD-Pseudo-Labeling}, and {\em SPD-Cross-Entropy}, respectively. For a fair comparison, we use the ``13-CNN'' architecture~\cite{tarvainen2017mean} for all methods across different datasets. We keep most hyperparameters introduced by the SSL methods close to their original papers~\cite{arazo2020pseudo,berthelot2019mixmatch,laine2016temporal}, since they can be easily integrated into our framework without massive adjustments. 


In Table~\ref{table:sslbackbones}, we list these four algorithms in the left column from weak to strong according to their performance in their original papers. The test accuracies demonstrate their corresponding performance for NLL based on our framework. {\em SPD-MixMatch} and {\em SPD-Pseudo-Labeling} outperform {\em SPD-Temporal Ensembling} by a large domain in both CIFAR-10 and CIFAR-100, especially under 80\% noise ratio (over 40\% in CIFAR-10). This phenomenon is reasonable because {\em Temporal Ensembling}~\cite{laine2016temporal} only uses consistency regularization for unsupervised loss, while {\em MixMatch}~\cite{berthelot2019mixmatch} and {\em Pseudo-Labeling}~\cite{arazo2020pseudo} also leverage entropy regularization as well as {\em MixUp} data augmentation~\cite{zhang2017MixUp}. Moreover, {\em SPD-Pseudo-Labeling} achieves remarkable test accuracy under 80\% noise ratio in CIFAR-100, which is +21.44\% higher than {\em SPD-MixMatch} and +42.66\% higher than {\em SPD-Temporal Ensembling}. We assume that this is due to the additional loss used in {\em SPD-Pseudo-Labeling} that prevents the model from assigning all labels to a single class at the early training stage. 

From the results of {\em SPD-Cross-Entropy}, we can see that after the removal of the SSL backbone, the test accuracy drops dramatically compared to {\em SPD-MixMatch} and {\em SPD-Pseudo-Labeling}, especially in high noise ratios and datasets with more classes (e.g., CIFAR-100). This is possibly due to the substantial amount of data that has been removed by the {\em self-prediction divider}, leaving very few samples per class. Thus, instead of discarding noisy samples, transferring them to unlabeled ones in SSL backbones is an effective way to combat noisy labels.


\begin{table}[!t]
\centering
\resizebox{0.40\textwidth}{!}{
\setlength{\tabcolsep}{5mm}{
\begin{tabular}{l|c}
\toprule
Methods & Test Accuracy \\
\midrule
Cross-Entropy &  69.21 \\
F-correction~\cite{patrini2017making} &  69.84 \\
M-correction~\cite{arazo2019unsupervised} &  71.00 \\
Joint-Optim~\cite{tanaka2018joint} &  72.16 \\
Dividemix~\cite{li2020DivideMix} &   73.91 \\
\midrule
GPL(ours) &  73.19 \\

Dividemix+(ours) &  $\boldsymbol{74.14}$  \\
\bottomrule

\end{tabular}
}
}
\setlength{\belowcaptionskip}{-0.6cm} 
\caption{Test accuracy (\%) on Clothing1M.}\label{table:clothing} 
\end{table}

\section{Conclusion}
This paper proposes a versatile framework called {\em SemiNLL} for NLL. This framework consists of two main parts: the mini-batch SS strategy and the SSL backbone. We conduct extensive experiments on benchmark-simulated and real-world datasets to demonstrate that {\em SemiNLL} can absorb a variety of SS strategies and SSL backbones, leveraging their power to achieve state-of-the-art performance in different noise scenarios. Moreover, we throughly analyze the effects of the two components in our framework. 

\bibliographystyle{plainnat}
\bibliography{arxiv-template}

\end{document}